%% file: root.tex
\documentclass[letterpaper, 10 pt, conference]{ieeeconf}  

\IEEEoverridecommandlockouts                              

\usepackage{graphicx}
\usepackage{amsmath}
\usepackage{amssymb}
\usepackage{booktabs}
\usepackage{bm}
\usepackage{times}
\usepackage{epsfig}
\usepackage{mathrsfs}
\usepackage{blindtext}
\usepackage[english]{babel}
\usepackage{color}
\usepackage[noend]{algpseudocode}
\usepackage{algorithmicx,algorithm}
\let\labelindent\relax
\usepackage[shortlabels]{enumitem}
\usepackage{multirow}
\usepackage{colortbl}
\usepackage{caption}
\usepackage{subcaption}
\usepackage{listings}
\usepackage{xcolor}
\input{math_commands.tex}

\title{\LARGE \bf
ASSIST: Interactive Scene Nodes \\for Scalable and Realistic Indoor Simulation
}

\begin{document}
\thispagestyle{empty}
\pagestyle{empty}

   

\author{Zhide Zhong$^{1,2*}$, Jiakai Cao$^{1,3*}$, Songen Gu$^{1,4}$, Sirui Xie$^{1,5}$, Weibo Gao$^{6}$, Liyi Luo$^{1,7}$,\\ Zike Yan$^{1,\dag}$, Hao Zhao$^{1}$, Guyue Zhou$^{1}$ 
\thanks{*Equal contribution, \dag  corresponding author}
\thanks{$^{1}$Institute for AI Industry Research (AIR), Tsinghua University, China,
    \{yanzike, zhaohao, zhouguyue\} @air.tsinghua.edu.cn.}%
\thanks{$^{2}$ Beijing Institute Of Technology, zhongzd@bit.edu.cn.}%
\thanks{$^{3}$ Institute of Computing Technology, Chinese Academy of Science, caojiakai21@mails.ucas.ac.cn
.}%
\thanks{$^{4}$ Institute of Software, Chinese Academy of Sciences, gusongen22@mails.ucas.ac.cn.}%
\thanks{$^{5}$ Beijing University of Civil Engineering and Architecture, 202105020134@stu.bucea.edu.cn}%
\thanks{$^{6}$ University of Science and Technology of China, weibogao@mail.ustc.edu.cn.}
\thanks{$^{7}$ McGill University, liyi.luo@mail.mcgill.ca
.}%
}

\makeatletter
\let\NAT@parse\undefined
\makeatother
\makeatletter
\g@addto@macro\@maketitle
{
  \begin{figure}[H]
  \setlength{\linewidth}{\textwidth}
  \setlength{\hsize}{\textwidth}
  \vspace{-4mm}
  \setcounter{figure}{0}  
  \centering
  \begin{tabular}{@{}c@{\hspace{1mm}}c@{}}
 	\includegraphics[width=0.99\linewidth]
        {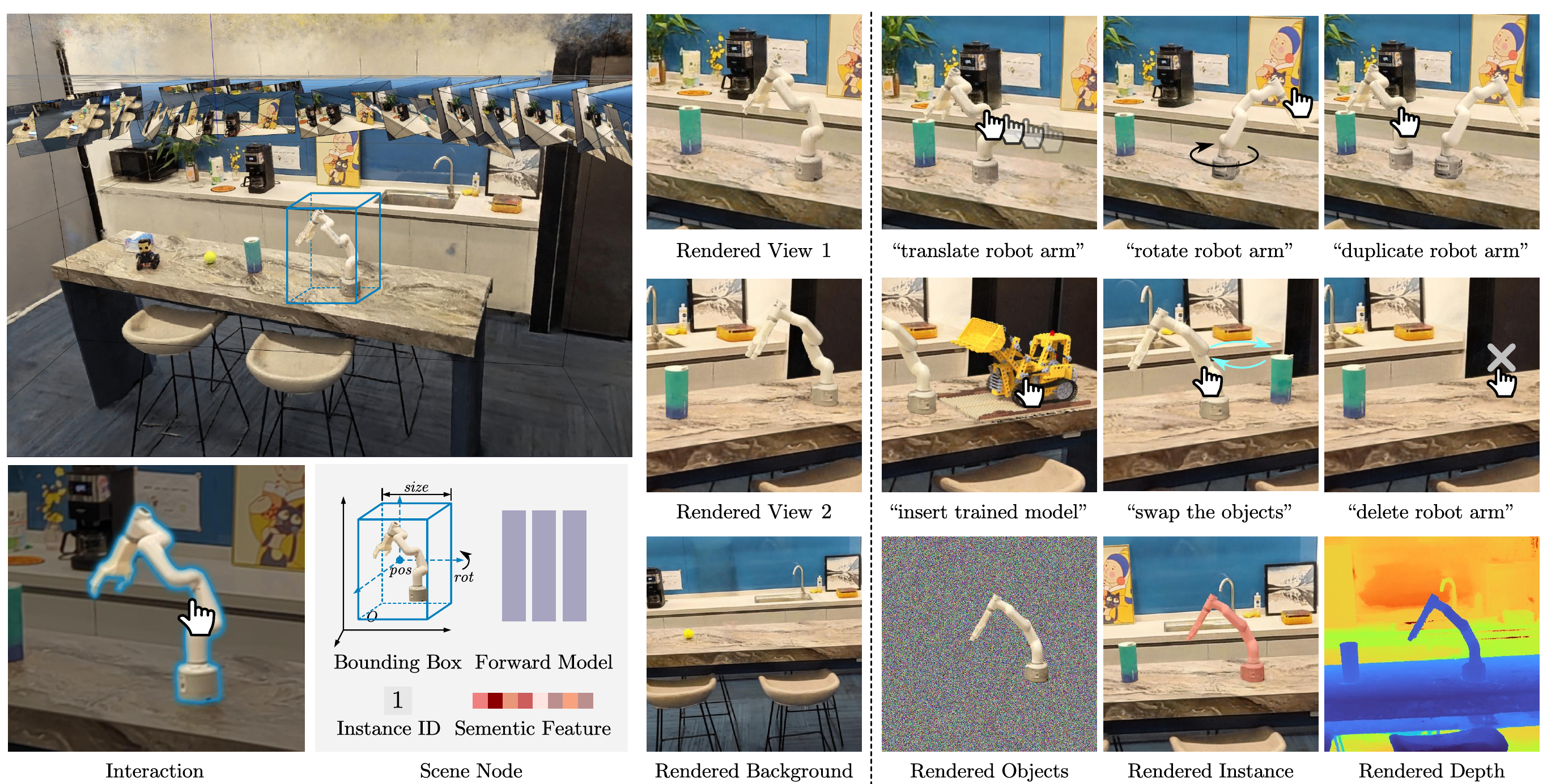}
  \end{tabular}
  \vspace{-2mm}
  \caption{We introduce a scalable and realistic neural simulator based on mouse click-and-drag or text controls. As illustrated in the figure, our input is a freely captured video. Users can interactively edit the scene to generate diverse new views, such as moving objects, rotating them, changing their size, removing them, or even inserting pre-trained object models from others. Furthermore, our approach can render individual object and background, depth maps, and highlight objects from new perspectives.
  }
  \label{fig:teaser}
  \vspace{-4mm}
  \end{figure}
}
\makeatother
\maketitle

\begin{abstract}
We present ASSIST, an object-wise neural radiance field as a panoptic representation for compositional and realistic simulation. Central to our approach is a novel scene node data structure that stores the information of each object in a unified fashion, allowing online interaction in both intra- and cross-scene settings. By incorporating a differentiable neural network along with the associated bounding box and semantic features, the proposed structure guarantees user-friendly interaction on independent objects to scale up novel view simulation. Objects in the scene can be queried, added, duplicated, deleted, transformed, or swapped simply through mouse/keyboard controls or language instructions. Experiments demonstrate the efficacy of the proposed method, where scaled realistic simulation can be achieved through interactive editing and compositional rendering, with color images, depth images, and panoptic segmentation masks generated in a 3D consistent manner.
\end{abstract}

\input{srcs/intro}

\input{srcs/related_work}

\input{srcs/method}

\input{srcs/experiments}

\section{CONCLUSION}
In summary, we present ASSIST, an object-wise neural rendering simulator equipped with interactive 3D scene editing capabilities. To achieve this, we decompose the complete scene into scene nodes, offering users the flexibility to generate editing commands via mouse actions or text descriptions. Through experiments on real and synthetic data, we validate the efficacy of our proposed method. We believe this approach holds significant promise for enhancing the quality and scale of 3D data. To achieve more realistic scene editing, exploring the effects of appearance embeddings across different scenes and integrating scene illumination models into the framework are promising directions in future work.


\bibliographystyle{IEEEtran}
\bibliography{reference}

\addtolength{\textheight}{-12cm}   





\end{document}

%% file: math_commands.tex
\usepackage{amsmath,amsfonts,bm}

\def\Figref#1{Fig.~\ref{#1}}

\def\Tabref#1{Tab.~\ref{#1}}

\def\eqref#1{eq.~\ref{#1}}
\def\Eqref#1{Eq.~\ref{#1}}

\def\1{\bm{1}}

\DeclareMathAlphabet{\mathsfit}{\encodingdefault}{\sfdefault}{m}{sl}
\SetMathAlphabet{\mathsfit}{bold}{\encodingdefault}{\sfdefault}{bx}{n}



%% file: srcs/intro.tex
\section{INTRODUCTION}
The recent trend in language and vision communities indicates that scaling and broadening the data is one simple but effective way to empower generalist models. Nevertheless, as robots perceive and interact in a 3D physical world, scaling up data collection for robot learning is non-trivial. Real-world data collection is usually laborious and task-specific~\cite{Brohan2022arxiv,Dhruv2023arxiv}, while synthetic scene simulators~\cite{Habitat2019iccv,Robothor2020cvpr,Li2022corl} struggle to recover high-fidelity models and reduce sim-to-real gap. Finding a versatile way to generate realistic data is of great importance toward scalable robot learning.

The introduction of Neural Radiance Field (NeRF)~\cite{Mildenhall2020eccv} allows differential rendering and guarantees multi-view consistency~\cite{Barron2021iccv, Barron2022cvpr}. The capability of photorealistic novel view synthesis shows great potential to bridge the sim-to-real gap~\cite{Yang2023cvpr,Byravan2023icra}. However, the scene-specific nature of NeRF hinders its application to scalable simulation: The 3D environment is inherently structured, and assets within the environment should be placed at any logical locations and reused for augmented data generation. This calls for an object-compositional structure of the neural radiance field, where the set of object models trained by individuals can serve as a library that best enforces data-sharing for simulation scalability.

In this paper, we present ASSIST, an object-compositional NeRF that enables user-friendly interaction. Key to our method is a scene node data structure that explicitly decomposes the scene into independent models. Bounded by a bounding box and represented as a differentiable neural network, each instance can be separately rendered and edited. Appearance embeddings and visual-language features are stored within the node structure for convenient exposure compensation and language queries. As illustrated in~\Figref{fig:teaser}, each scene is cast as a panoptic representation that allows for online scene editing via user-friendly interactions. The structures of different scenes are maintained in a unified manner that supports intra- and cross-scene operations such as query, addition, replication, translation, rotation, deletion, and inter-object swapping through mouse/keyboard controls or using language instructions. We show in the experiments that realistic rendering given novel views and novel object organizations can be generated online without retraining. The key contributions of the proposed method can be categorized as follows:

$\bullet$ We introduce a scene node structure that explicitly decomposes the scene at an object level.

$\bullet$ We exploit the scene node structure to achieve object-wise scene editing through convenient online interaction.

$\bullet$ We conduct compositional rendering with the combination of scene nodes towards scalable and photorealistic simulation.

%% file: srcs/related_work.tex











\section{RELATED WORK}
\subsection{Scaling Robot Learning}
"Scaling robot learning"~\cite{Yu2023rss} refers to boosting robotics generalization ability through learning from large-scale data. Achieving the scalability of robot learning includes multi-robot deployment for data collection, generating vast amounts of training data through simulation, or integrating human daily feedback into the learning process. RT-2~\cite{brohan2023rt}, for instance, learns a visual language action model (VLA) from the web and robotic data, enabling robot control through generalist instructions. Gathering the dataset at scale is extremely costly. ROSIE~\cite{Yu2023rss} employs text-to-image diffusion models for proactive data augmentation, where text guides the restoration of various unseen objects or backgrounds. Similarly, GenAug~\cite{chen2023genaug} "augments" RGBD images through text-image generative models, resulting in a significant improvement in generalizing to novel scenes. DALL-E-Bot~\cite{kapelyukh2023dall} enables a robot to rearrange objects in a scene, by first inferring a text description of those objects, then generating an image representing a natural, human-like arrangement of those objects, and finally physically arranging the objects according to that goal image. We share a similar idea of augmenting data through object insertion and rearrangement, but seek a realistic simulation with user-friendly interaction through object-compositional neural radiance fields.

\subsection{Object-compositional Neural Radiance Field}
Photorealistic simulation environment seems another route toward scaling and broadening robotic data~\cite{Li2022corl}. Recent advances in learning of neural radiance field show that building an editable digital twin of the real-world environment can be applied for building safe autonomy systems~\cite{Yang2023cvpr}. Similar efforts have been made~\cite{Ost2021cvpr, Kundu2022cvpr, Wu2023CICAI} to decompose dynamic scenes with multiple moving cars into a learned scene graph. Apart from the autonomous driving scenarios, the concurrent works of~\cite{Kong2023cvpr, Han2023ral} introduce object-decomposed NeRF-SLAM systems. Different object models are trained simultaneously to accelerate convergence and reduce the total parameters. 
In contrast to the explicit decomposition of the global neural radiance field, there are also implicit decomposition manners that output the semantic/instance channels with a shared global network~\cite{Wang2023iclr,Yang2021iccv, Wu2022eccv, Wang2023iccv}. Although the shared weights bring semantic consistency across views and against noise~\cite{Zhi2021piccv, Siddiqui2023cvpr}, the training and inference of different objects share the entire network to get the shape of a desired object.

\begin{figure*}[tb]
      \centering
      \includegraphics[width=0.99\linewidth]{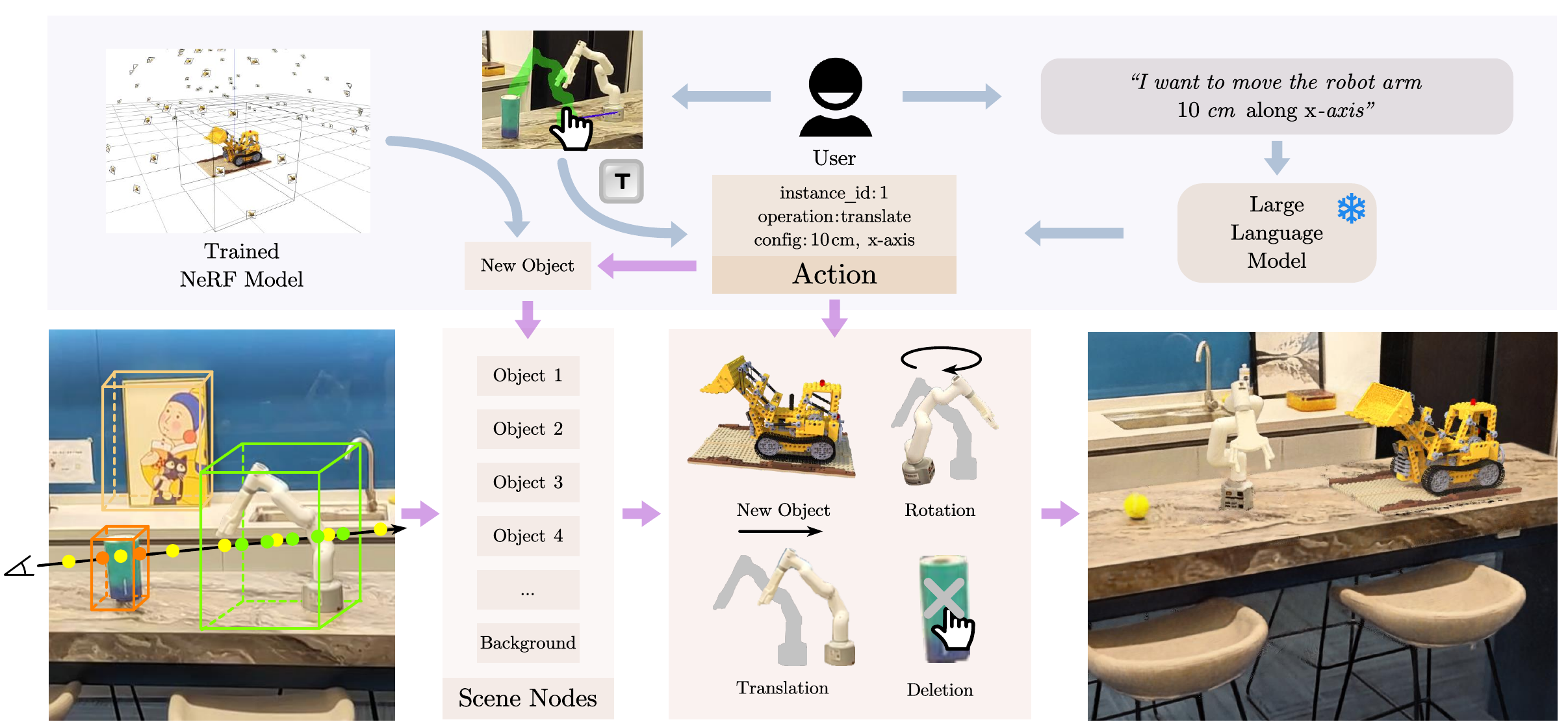}
      \caption{We organize the entire scene using individual scene nodes. Users can generate action commands for object nodes either by mouse dragging combined with keyboard or through text controls. These commands include actions like moving (translate and rotate), scaling, or inserting new objects into the scene. Ultimately, these action commands are applied to the nodes to achieve the scene editing.}
      \label{fig:overview}
\end{figure*}

\subsection{Editing a Trained NeRF}
In the field of computer vision, traditional data augmentation techniques are commonly used to introduce variations to original images. These variations, such as changes in appearance or position, help improve the diversity of the training data and enhance the generalization ability of models. In the context of NeRF-based 3D assets, there is a growing need for user-friendly interactive editing, enabling efficient generation of diverse models.

To address this need, several recent works have introduced methods for interactive editing of NeRF-based scenes, so diverse models can be generated efficiently. Instruct Nerf2nerf~\cite{Haque2023iccv} introduces a method to iteratively edit the input images and optimize the underlying scene through re-trainng, resulting in a new 3D scene that aligns with the text instructions. Seal-3D~\cite{Wang2023iccvseal} presents the first interactive pixel-level editing system, offering fine-grained editing tools for both geometry and color. Nerfshop~\cite{Jambon2023nerfshop} constructs a tetrahedral mesh. The data structure of the radiance field allows users to manipulate the mesh vertices and enable free-form deformations and edits. Our objective is to target user-friendly interaction for arbitrary object placement in NeRF-based scenes. By enabling such interaction, we aim to scale simulations with diverse organization of assets, allowing users to easily arrange objects and generate multi-view consistent data according to their requirements.


%% file: srcs/method.tex
\section{METHOD}
The proposed method takes multi-view images as input, and outputs an interactive simulation environment for generating photorealistic and 3D consistent data.
\subsection{Interactive Scene Nodes}
We represent the scene as a panoptic neural radiance field, where the model of each object is an independent scene node. The node structure stores the object's instance ID, semantic features using CLIP embeddings~\cite{Radford2021icml}, a forward model for differential rendering, and the associated 3D bounding box. Each node can be interactively edited through mouse/keyboard controls or language instructions. 

\definecolor{codegreen}{rgb}{0,0.6,0}
\definecolor{codegray}{rgb}{0.5,0.5,0.5}
\definecolor{codepurple}{rgb}{0.58,0,0.82}
\definecolor{backcolour}{rgb}{0.95,0.95,0.92}
\lstset{
  backgroundcolor=\color{backcolour},
  commentstyle=\color{codegreen},
  keywordstyle=\color{magenta},
  numberstyle=\tiny\color{codegray},
  stringstyle=\color{codepurple},
  basicstyle=\normalsize\ttfamily,
  breakatwhitespace=false,     
  breaklines=true,                 
  captionpos=b,                    
  keepspaces=true,                 
  numbers=none,                    
  numbersep=10pt,                  
  showspaces=false,                
  showstringspaces=false,
  showtabs=false,                  
  tabsize=2,
  language=C++,
}

\begin{lstlisting}
struct scene_node{
    int id;
    float embedding[512];
    struct forward_model;
    struct bbox;
}
\end{lstlisting}

As illustrated in~\Figref{fig:overview}, the interaction is converted to a standard structure that defines the target object, the specific operation, and the corresponding configurations. For example, we can simply enter the instruction of "move the robot arm 10 cm along x-axis" or use mouse control to drag the robot arm directly. The interaction will be then transformed into a structure of \{instance\_id: 1; operation: translate; config: 10 cm, x-axis\} that acts on the corresponding scene node. Note that the bounding box defines the local canonical space of the object. Interactions that define object pose and placement can be operated directly on the bounding box and transform the object to the target coordinate.

In practice, for instruction interaction, we use a large language model (GPT-4~\cite{Openai2023gpt} from OpenAI API) to split the high-level instruction into the pre-defined action structure through question-answering. In terms of mouse/keyboard actions, we choose the target operation through keyboard control and conduct the specific operation on the mouse-clicked object. 
The proposed simulator supports the following operations that boost convenient data augmentation in the three-dimensional space:

$\bullet$ Translate (\texttt{t}).
With the defined canonical space and the configuration of \{axis, distance\}, we can move the object along the axis given the input distance value.

$\bullet$ Rotate (\texttt{r}).
With the defined canonical space and the configuration of \{axis, angle\}, we can scale the object along the axis given the input value.

$\bullet$ Scale (\texttt{x}).
With the defined canonical space and the configuration of \{axis, scale\}, we can rotate the object along the axis given the input rotation value.

$\bullet$ Select (\texttt{SPACE}).
With the selected node, the instance mask can be highlighted by adding a specific RGB value to the ray samples within the bounding box.

$\bullet$ Replicate (\texttt{+}). 
Replication can be achieved simply by instantiating a new scene node and copying the corresponding node values. 

$\bullet$ Delete (\texttt{-}). 
Deleting an object is simply a removal operation on the target node from the scene node list.

$\bullet$ Cross-scene operation (\texttt{c}). 
As the scene node structure is shared among different trained models, we can easily achieve cross-scene operations such as adding (\texttt{a}) or swapping (\texttt{s}) by first loading the second panoptic neural radiance field and the associated scene nodes and then performing the above-mentioned operations.

\subsection{Compositional Rendering with Scene Nodes}
The bottom left corner of \Figref{fig:overview} demonstrates our compositional rendering process. Given a camera pose $\mathcal{T}_i$, the ray $\mathbf{r} = \mathbf{o} + t\mathbf{d}$ emitted from the optical center $\mathbf{o}$ through a pixel can be determined. For the background node and all object nodes the ray emits through, we sample $N$ points separately. That is to say, we can separate the whole ray into $m+1$ parts of $\mathbf{r} = \{\mathbf{r}^{\text{bg}}, \mathbf{r}^{\text{obj1}}, \cdots, \mathbf{r}^{\text{objm}}\}$, where $m$ is the number of intersected objects. As the intersection of the ray and the bounding box determines the starting and ending points $\{\mathbf{r}(t_\text{in}),\mathbf{r}(t_\text{out})\}$ of the ray through each object, points inside each object's bounding box are then transformed from the world space to their local canonical space accordingly. 

The color and density values of the sampled points can be queried through forward passes given the associated forward models. By sorting the samples according to their depth values as $P_i\in\mathtt{sorted}(\{P_i^{\text{bg}}, P_i^{\text{obj1}},\cdots,P_i^{\text{objm}}\})$, the rendering can be achieved through a composition of the standard volume rendering formula~\cite{Mildenhall2020eccv} as: 
\begin{equation}
\hat{C}(\boldsymbol{r}) = \sum_{P_i}T_i\alpha_i {c}_i,\ \ T_i = \exp(-\sum_{k=1}^{i-1}\sigma_k\delta_k),
\label{eq:compositinal_rgb}
\end{equation}
where $\alpha_i=1-\exp(-\sigma_i\delta_i), \delta_i=t_{i+1}-t_i$. 

In addition to rendering the color of the entire ray, we can also perform volume rendering separately for each node to generate the pixel color, depth and opacity as:
\begin{equation}
\hat{C}^k(\boldsymbol{r}^k)=\sum_{i=1}^N T_i^k\left(1-\exp \left(-\sigma_i^k \delta_i^k\right)\right) \mathbf{c}_i^k,
\label{eq:object_rgb}
\end{equation}
\begin{equation}
\hat{D}^k(\boldsymbol{r}^k)=\sum_{i=1}^N T_i^k\left(1-\exp \left(-\sigma_i^k \delta_i^k\right)\right)t_i^k,
\label{eq:object_depth}
\end{equation}
\begin{equation}
\hat{O}^k(\boldsymbol{r}^k)=\sum_{i=1}^N T_i^k\left(1-\exp \left(-\sigma_i^k \delta_i^k\right)\right).
\label{eq:object_accumulation}
\end{equation}
where the background node is indexed as $0$;  $T_i^k=\exp \left(-\sum_{j=1}^{i-1} \sigma_i^k \delta_j^k\right)$ is the accumulated transmittance along the ray for the k-th object.

\subsection{Optimization of Node Models}
We aim to recover an object-compositional neural radiance field, where the rendering of each node can be conducted accurately. The photometric error minimization regarding the compositional rendering equals to the training of a single vanilla NeRF as:
\begin{equation}
\mathcal{L}_{\text {comp\_rgb}}= \left\|\hat{C}(\boldsymbol{r})-C(\boldsymbol{r})\right\|_2^2.
\label{eq:comp_rgb}
\end{equation}

Note that all samples along the ray contribute to the final pixel color value. The gradient through back-propagated errors will update the forward models for all nodes, thus leading to the decomposition ambiguity. The surface point of an object may exhibit low density, and the color of the corresponding pixel may be contributed mainly by the background node. To enforce the gradient contribution to the right object node, we employ an object accumulation loss and an object color loss to supervise the associated object models. Given a binary mask $M^k(\boldsymbol{r^k})$ that indicates if the ray first hits the associated object surface, the accumulation of each object should be consistent with the pixel-node association, and the true object model should render the corresponding pixel color:
\begin{equation}
\mathcal{L}_{\text {obj\_acc}}=\sum_{k=1}^{K} \left\|\hat{O}^k(\boldsymbol{r}^k)-M^k(\boldsymbol{r}^k)\right\|_2^2,
\label{eq:obj_acc}
\end{equation}
\begin{equation}
\mathcal{L}_{\text {obj\_rgb}}=\sum_{k=1}^{K} M^k(\boldsymbol{r}^k) \left\|\hat{C}^k(\boldsymbol{r}^k)-C(\boldsymbol{r})\right\|_2^2.
\label{eq:obj_rgb}
\end{equation}

The losses of~\Eqref{eq:obj_acc} and~\ref{eq:obj_rgb} enforce the attribution of pixel color to the associated node model, and avoid the generation of floaters when a ray traverses an object-bounding box without hitting the surface. However, the areas occluded by the objects lacks proper supervision. Imagine a book on a table. The areas of the table occluded by the book does not provide any supervisory signals to the model. As a result, the rendered color of the area is ambiguous. This significantly impacts the editing results when we move the foreground object away. To address the issue, we use a pre-trained inpainting model of LAMA~\cite{suvorov2022resolution} to predict the appearance of the occluded areas and supervise the color rendered by the background model. As LAMA is inherently a 2D model that lacks multi-view consistency, we also adopt a monocular depth estimation network~\cite{eftekhar2021omnidata} to predict the depth map of the inpainted image and serves as an auxiliary supervisory signal~\cite{yu2022monosdf}:
\begin{equation}
\mathcal{L}_{\text {bg\_rgb}}= \left\|\hat{C}^0(\boldsymbol{r})-C_{\text {inpaint}}(\boldsymbol{r})\right\|_2^2
\end{equation}
\begin{equation}
\mathcal{L}_\text{bg\_depth} = \left\| w\hat{D}^0(\textbf{r}) + q -D_{\text {inpaint}}(\boldsymbol{r}) \right\|_2^2,
\end{equation}
where $w$ and $q$ are learnable scale and shift factors that are optimized through training for scale and shift invariance.

The final loss for training our panoptic neural radiance field consists of five terms:
\begin{equation}
\begin{aligned}
\mathcal{L} = &\mathcal{L}_{\text {comp\_rgb }}+\lambda_1 \mathcal{L}_{\text {bg\_rgb }}+\lambda_2 \mathcal{L}_{\text {bg\_depth }} \\ &+\lambda_3 \mathcal{L}_{\text {obj\_acc }}+\lambda_4 \mathcal{L}_{\text {obj\_rgb}},
\end{aligned}
\end{equation}
where $\lambda_1, \lambda_2, \lambda_3, \lambda_4$ are normalizing constants to balance the training.

%% file: srcs/experiments.tex
\section{EXPERIMENTS}
We conduct thorough experiments on multiple scenes qualitatively and quantitatively. Comparisons against state-of-the-art methods along with detailed ablation studies are provided. Further results concerning the user-friendly editing and photorealistic novel view synthesis justify the efficacy of the proposed method.

\subsection{Experimental Setups}
\noindent\textbf{Datasets}
We train our panoptic scene representation on multiple datasets. For quantitative evaluation, we follow the experimental setting of UDC-NeRF~\cite{Wang2023iccv} and evaluate the proposed method on ToyDesk~\cite{Yang2021iccv} and ScanNet~\cite{Dai2017cvpr} datasets. We also capture an inward-facing indoor scene using multi-view images from a handheld phone with resolution $1280\times720$. The camera poses are calculated using COLMAP~\cite{colmap}, while the instance masks for supervising foreground nodes are estimated using XMem~\cite{xmem}. We also take the commonly used synthetic scenes~\cite{Mildenhall2020eccv} of Lego and Mic for cross-scene evaluation.

\noindent\textbf{Implementation Details}
We take the default model (Nerfacto) of NeRFStudio~\cite{Tancik2023siggraph} as our forward model architecture among scene nodes. All experiments were conducted on a server with an AMD EPYC 7742 64-Core Processor and an NVIDIA GeForce RTX A4000 graphics card. Each model is trained for $30,000$ iterations using the Adam optimizer at a learning rate of $1e\text{-}2$. An exponential decay scheduler is applied, adjusting the learning rate to a final value of $1e\text{-}4$.  We also applied the proposal sampler strategy~\cite{Tancik2023siggraph} for each node to better represent the scene.

\begin{figure}[tb]
      \centering
      \includegraphics[width=0.98\linewidth]{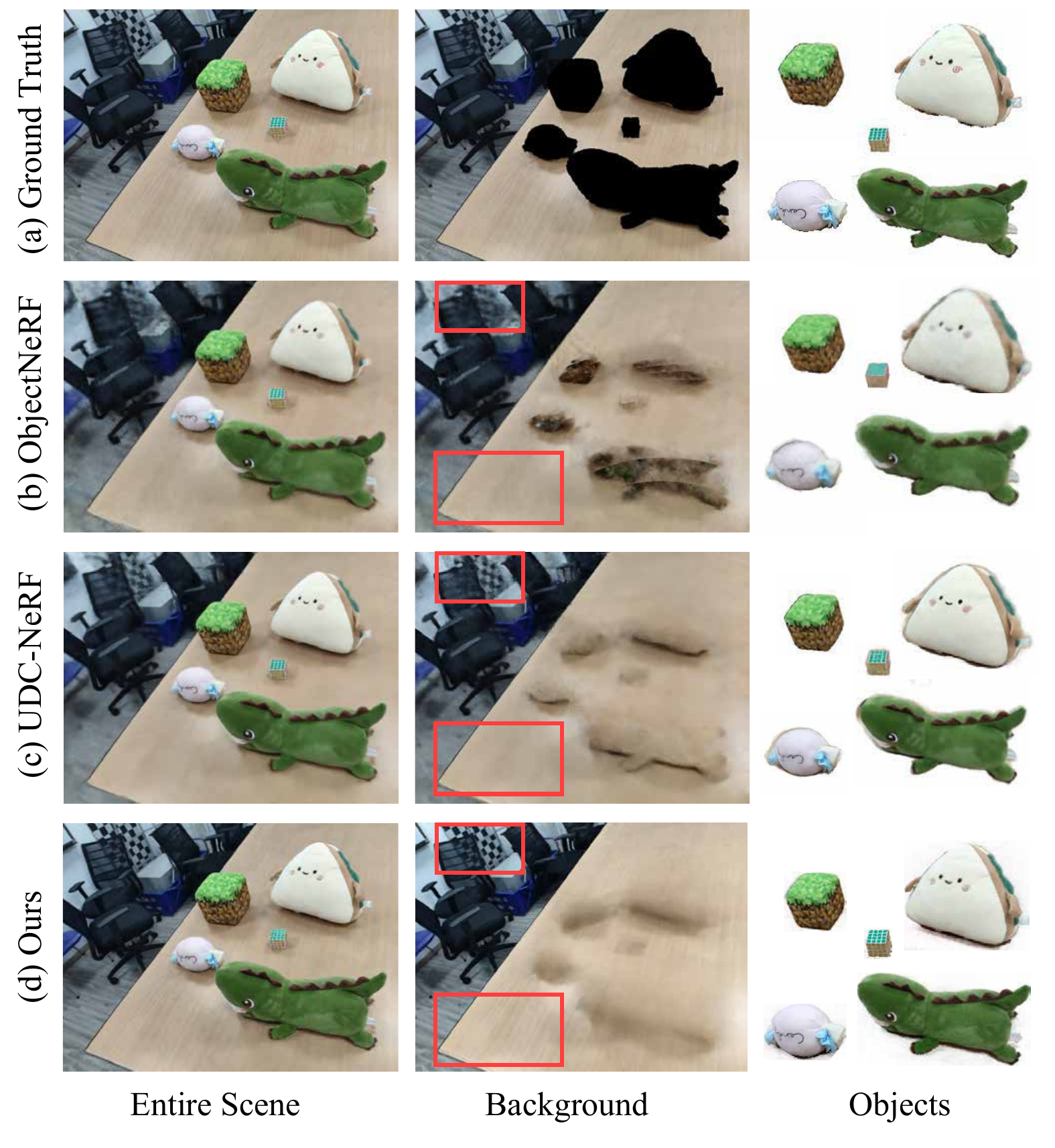}
      \caption{Results of the decomposition on the ToyDesk dataset are presented, in comparison with Object-NeRF \cite{Yang2021iccv} and UDC-NeRF \cite{Wang2023iccv}. We display the rendered ouputs from our composite radiance field (shown in the first column) alongside individual radiance fields for the background and objects (depicted in the second and third columns).}
      \label{fig:experiment_compare}
\end{figure}

\begin{table}[tb]
\caption{Quantitative comparison with other methods. The best results are shown in \textbf{bold}.}
\centering
\begin{tabular}{l|lccc}
\toprule
\textbf{Scenes} & \textbf{Methods} & \textbf{PSNR} $\uparrow$ & \textbf{SSIM} $\uparrow$ & \textbf{LPIPS} $\downarrow$ \\
\midrule
\multirow{3}{*}{ToyDesk2} & Object-NeRF~\cite{Yang2021iccv} & 24.815 & 0.7888 & 0.446 \\
& UDC-NeRF~\cite{Wang2023iccv} & 25.756 & 0.8126 & 0.448 \\
& Ours & \textbf{26.552} & \textbf{0.8507} & \textbf{0.186} \\
\midrule
\multirow{3}{*}{ScanNet} & Object-NeRF~\cite{Yang2021iccv} & 25.264 & 0.8047 & 0.4094 \\
& UDC-NeRF~\cite{Wang2023iccv} & 26.135 & 0.8249 & 0.395 \\
& Ours & \textbf{30.360} & \textbf{0.8394} & \textbf{0.236} \\
\bottomrule
\end{tabular}
\label{tab:compare}
\end{table}

\begin{figure*}
      \centering
      \includegraphics[width=0.99\linewidth]{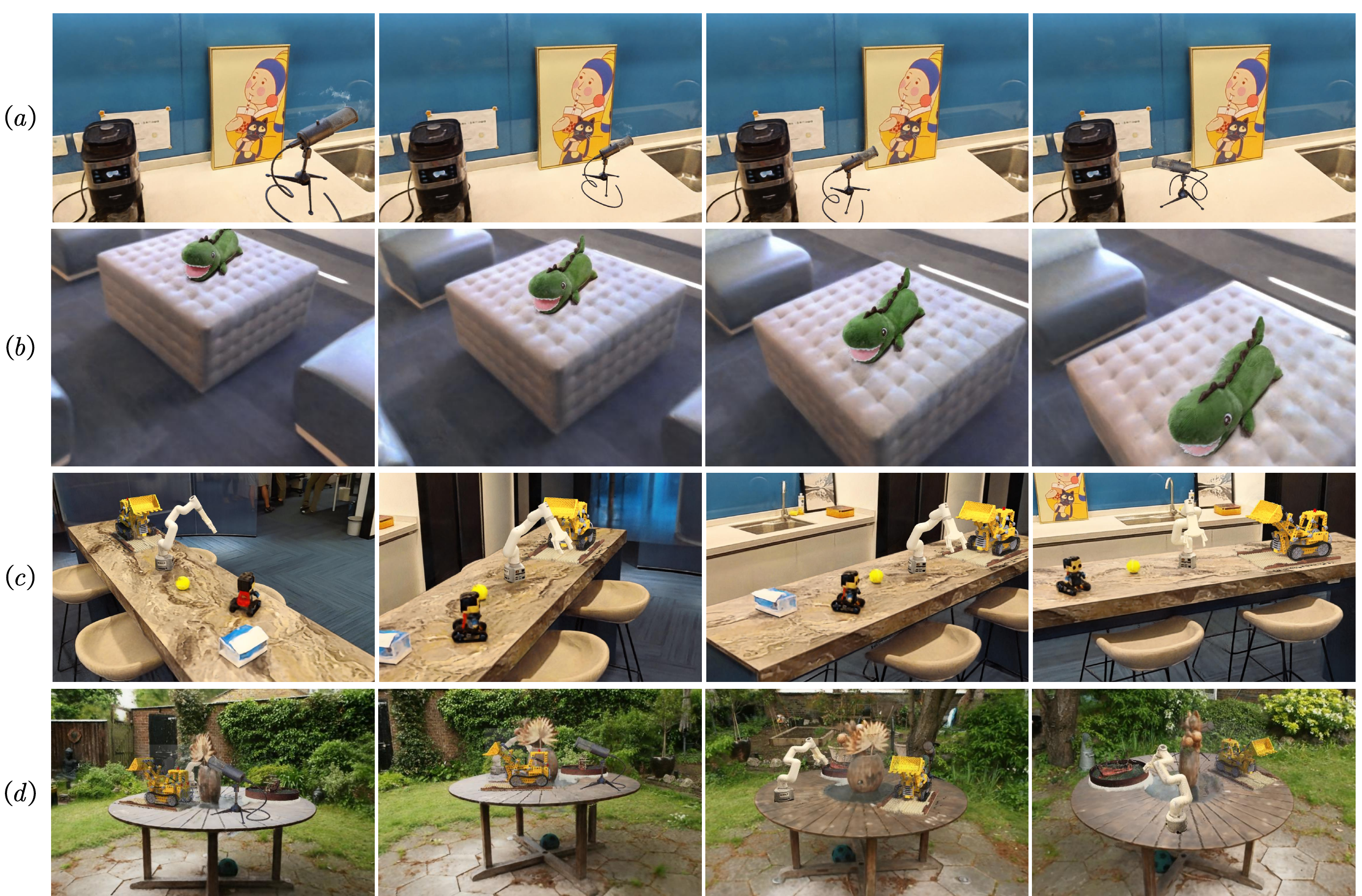}
      \caption{The qualitative results of cascaded editing (a) and cross-scene editing (in four views, b-d): (a) Add pretrained Mic model and apply scaling-translation-rotation cascadingly; (b) Add Dragon from ToyDesk to ScanNet; (c) Add Lego to our real-captured scene; (d) Add multiple objects to Mip-NeRF 360 dataset.}
      \label{fig:experiment_editing}
\end{figure*}

\subsection{Comparison against Baselines}
We conduct experiments to evaluate the synthesized novel views of both foreground nodes and the background node. The results are compared against the SOTA methods of Object-NeRF~\cite{Yang2021iccv} and UDC-NeRF~\cite{Wang2023iccv}. As illustrated in~\Figref{fig:experiment_compare}, the proposed method yields the best completion results. For the background node, the black shadows are effectively removed while recovering fine-grained details. In terms of object decomposition, the proposed method provides more precise edges with clear details compared to other methods. The quantitative results shown in~\Tabref{tab:compare} further prove the effectiveness of the proposed method. Thanks to the explicit decomposition of the independent scene nodes, the sampling strategy better fits the local canonical space. Without weight sharing between semantic and RGB heads, the potential interference is avoided with promising accuracy.

\subsection{Ablation Studies}
\begin{figure}[tb]
      \centering
      \includegraphics[width=0.99\linewidth]{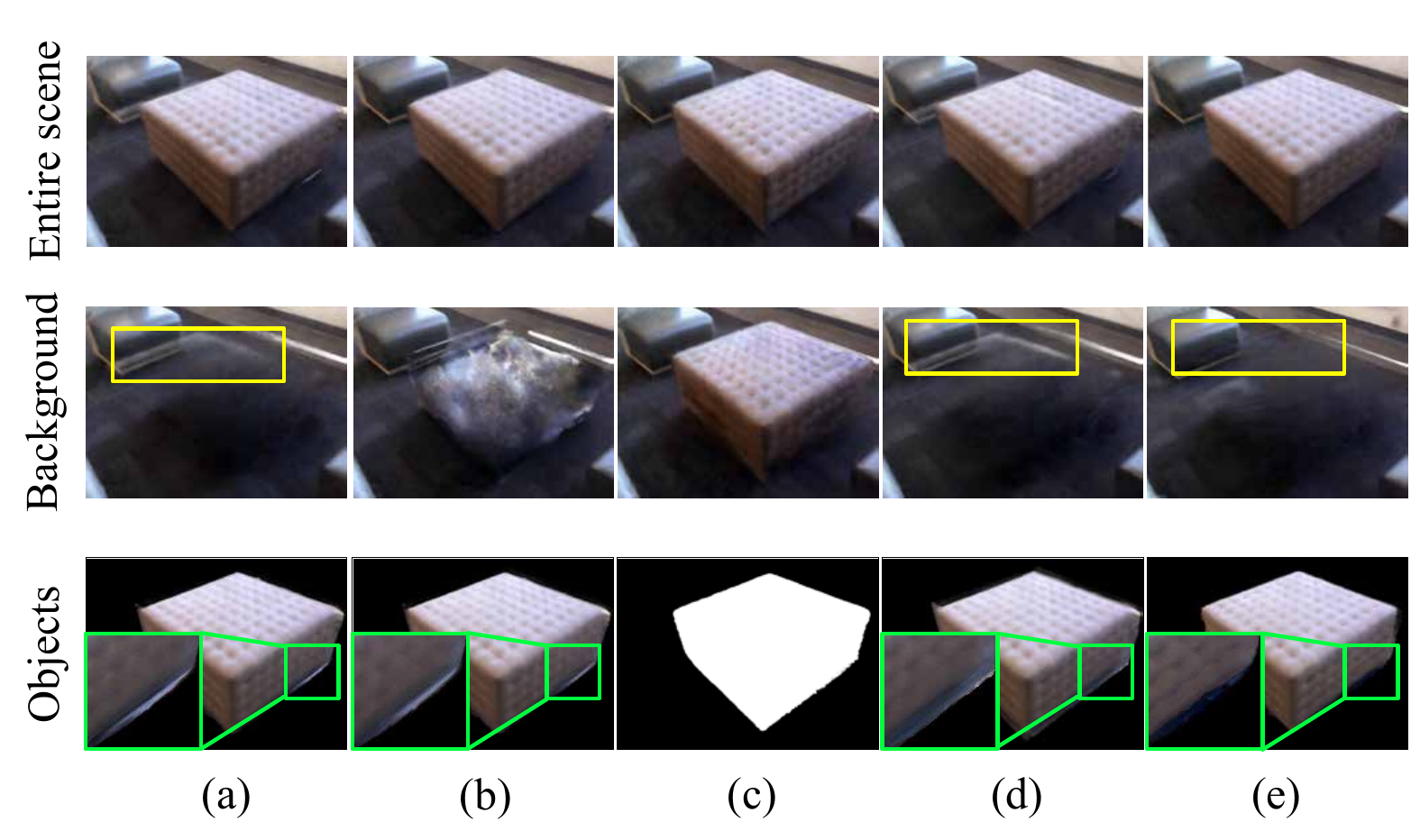}
      \caption{Qualitative results of the ablation study: (a) w/o. 2D inpainted pseudo depth supervision; (b) w/o 2D inpainted pseudo RGBD supervision; (c) w/o object RGB supervision and 2D inpainted pseudo RGBD supervision; (d) w/o object accumulation supervision. (e) Ours.}
      \label{fig:experiment_ablation}
\end{figure}

\begin{table}[tb]
\centering
\caption{Ablation studies on the ScanNet dataset. Best shown in \textbf{bold} and the second-best shown \underline{underlined}.}
\begin{tabular}{lccc}
\toprule
\textbf{Methods}    & \textbf{PSNR} $\uparrow$ & \textbf{SSIM} $\uparrow$ & \textbf{LPIPS} $\downarrow$ \\
\midrule
w/o Inpaint-Depth  & 30.68                   & 0.8852                   & 0.2127                       \\
w/o Inpaint-RGBD   & \underline{31.69}       & \textbf{0.8897}          & \textbf{0.2019}              \\
w/o Obj. RGB and Inpaint-RGBD  & 31.31       & 0.8746                   & 0.2182                       \\
w/o Obj. Acc.      & 30.99                   & 0.8886                   & 0.2110                       \\
Ours     & \textbf{31.95}          & \underline{0.8895}       & \underline{0.2057}           \\
\bottomrule
\end{tabular}%

\label{tab:ablation_toydesk2}
\end{table}

We also conducted detailed ablation studies on scene $0113\_00$ in ScanNet dataset to justify the design of our loss terms. As shown in~\Figref{fig:experiment_ablation} and~\Tabref{tab:ablation_toydesk2}, the role of each term is evident. From~\Figref{fig:experiment_ablation} (a) to (c), we can see that 2D inpainting can not only reduce the decomposition ambiguity, but lead to a clean foreground node without floaters. It can be seen in~\Tabref{tab:ablation_toydesk2} even though inpainting does not affect the compositional rendering as the occluded areas can not be seen, the model will fail to decompose the space into object and background node correctly. We can also see from~\Figref{fig:experiment_ablation} and~\Tabref{tab:ablation_toydesk2} that incorporating the accumulation loss results in cleaner areas in near-surface areas. Rays that traverse the foreground bounding box without hitting the object surface will not contribute to the integration of pixel color, thus avoiding the gradient back-propagation to the node models.



\subsection{User-friendly Interaction}

In this section, we visualize the results of user-friendly interactions, including operations such as duplication, swapping, deletion, translation, rotation, scaling, cross-scene editing and cascaded editing. Interactions are conducted through mouse controlling and language instruction. User can either click on objects within a 2D image or describe his instruction, then the selected object is highlighted both in the 2D view and the 3D renderer. Following this, users can choose various interactions, and they can set the next operation to the object by dragging the mouse or further text description. In \Figref{fig:experiment_editing}, we present the results of cross-scene editing operations performed on a real-captured scene, as well as cascaded editing operations to offer a qualitative representation. We have trained representative NeRF models from the NeRF dataset, such as Lego and Mic, and integrated them into our real-captured scenes. Furthermore, we demonstrate the outcomes of introducing multiple new objects into the Mip-NeRF 360 scene \cite{Barron2022cvpr}. These results underscore the adaptability and efficacy of our approach.


